%% file: root.tex
\documentclass[letterpaper, 10 pt, conference]{ieeeconf}  

\IEEEoverridecommandlockouts                              

\pdfoutput=1

\usepackage{graphicx} 
\usepackage{amsmath} 
\usepackage{siunitx}
\usepackage{url}
\usepackage{booktabs}
\usepackage{cite}
\usepackage{color}
\usepackage[hidelinks]{hyperref}

\usepackage[capitalise]{cleveref}
\usepackage[export]{adjustbox}

\newcommand{\daggeralg}{\textsc{DAgger}}

\crefname{equation}{}{} 
\crefname{section}{Sec.}{Sec.}

\title{\LARGE \bf
Learning to Compensate Photovoltaic Power Fluctuations \\from Images of the Sky by Imitating an Optimal Policy
}

\author{Robin Spiess, Felix Berkenkamp, Jan Poland, and Andreas Krause
\thanks{Felix Berkenkamp, and Andreas Krause are with the Learning \& Adaptive
	Systems Group (LAS), Department of Computer Science, ETH Zurich,
	Switzerland. Email: {\tt \{befelix, krausea\}@inf.ethz.ch}} %
\thanks{Robin Spiess is with the Department of Computer Science, ETH Zurich, Switzerland. Email: {\tt robin.spiess@alumni.ethz.ch}} %
\thanks{Jan Poland is with ABB Corporate Research, Switzerland. Email: {\tt jan.poland@ch.abb.com}} %
\thanks{This work was supported by SNSF grant {200020\_159557}, the Vector Institute, and an Open Philantropy Project AI fellowship.}
}

\begin{document}

\maketitle
\thispagestyle{empty}
\pagestyle{empty}

\input{sections/0-abstract}
\input{sections/1-introduction}
\input{sections/2-problem}
\input{sections/3-background}
\input{sections/4-control}
\input{sections/5-results}
\input{sections/6-conclusions}



\bibliographystyle{IEEEtran}
\bibliography{IEEEabrv,bibliography}

\end{document}

%% file: sections/0-abstract.tex

\begin{abstract}%

The energy output of photovoltaic (PV) power plants depends on the environment and thus fluctuates over time. As a result, PV power can cause instability in the power grid, in particular when increasingly used.
Limiting the rate of change of the power output is a common way to mitigate these fluctuations, often with the help of large batteries. A reactive controller that uses these batteries to compensate ramps works in practice, but causes stress on the battery due to a high energy throughput.
%
In this paper, we present a deep learning approach that uses images of the sky to compensate power fluctuations predictively and reduces battery stress.
%
%
In particular, we show that the optimal control policy can be computed using information that is only available in hindsight. Based on this, we use imitation learning to train a neural network that approximates this hindsight-optimal policy, but uses only currently available sky images and sensor data.
We evaluate our method on a large dataset of measurements and images from a real power plant and show that the trained policy reduces stress on the battery. 
\end{abstract}

%% file: sections/1-introduction.tex

\section{Introduction}

Photovoltaic (PV) power generation has grown at a rate of roughly 30\% per year in recent years and reached a global capacity of over \SI{400}{\giga\watt} at the end of 2017~\cite{iea2017snapshot}.
However, its fluctuations are known to negatively affect grid stability and may cause blackouts in the worst case.
To mitigate this problem, grid operators in various countries impose ramp rate limits on PV power plants \cite{gevorgian2013review, denmark2015}.
A common way to implement the ramp rate limitation is to use large batteries to compensate any shortfall between generated power and the output level enforced by the rate constraints.
However, repeatedly charging and discharging the batteries reduces their lifespan~\cite{vetter2005}.
Hence, the power plant owner is interested in minimizing the energy throughput of the battery to reduce costs.
If good information on the near future evolution of the PV power were available, the controller for the power output could make use of this and reduce the stress on the battery.

PV fluctuations arise from clouds moving between the sun and the PV panels.
In this paper, we design a policy to control the PV power output based on currently available sky images and sensor data. We realize this policy as a deep neural network and use imitation learning~\cite{ross2011dagger} to approximate an optimal reference policy that is given perfect knowledge about the PV power evolution.
We evaluate our method on a dataset of real-world images and measurements from a solar power plant and show that it improves performance over
the baseline reactive policy, which compensates steep PV ramps with the battery in an ad-hoc manner. Our main contribution is
the end-to-end learning of a control policy that acts on image input, by extracting the relevant information about the short-term future
from these images.

\subsection{Related work}
Most of the research about power production forecasts for solar power plants has worked with longer time horizons than what is required for our problem.
These predictions often use machine learning methods, such as neural networks, together with past data in form of past weather data~\cite{mellit2010},
numerical weather predictions~\cite{gensler2016forecastdeeplearning, leva2017analysis}, or satellite images~\cite{hammer1999short}.
However, these predictions are not useful to compensate short-term fluctuations. Instead, approaches that create ultra short-term forecasts of up to one hour into the future often utilize local sensor data and images of the sky. For example, the global horizontal irradiance (GHI) can be predicted using handcrafted features extracted from images~\cite{chu2015real} or cloud motion and cloud map forecasts~\cite{magnone2017cloudmotion, schmidt2017short}. The latter data can also be used to estimate shadow maps on the ground, which can be used to improve predictions of the power output~\cite{chow2011}.
Other work could serve as a basis for power output predictions. For example,~\cite{bernecker2013representation} classifies clouds while~\cite{pothineni2018} uses a neural network to predict whether the sky is clouded in five or ten minutes. All these approaches measure prediction quality using the prediction error, not control performance.

Despite the significant research conducted, short-term predictions are never perfect and their full value in the context of ramp rate limitation is not clear. In fact, when using these predictions
inside a model predictive control (MPC) approach, \cite{habicht2017PVRL} demonstrates that even small errors can compound, which results in poor performance on real world data. That work also studies deep reinforcement learning approaches
to our problem. While this performs well with artificially generated sky images, it turns out to be challenging on real world
data and does not yield an improvement over a reactive policy.

Another way to learn control policies is imitation learning, which uses supervised learning to train the policy. Imitation learning algorithms generate training data during the learning process, by labelling states visited by the trained policy with the optimal actions of a reference policy~\cite{daume2009search, ross2011dagger}. Imitation learning
has been used in several applications that include autonomous flight \cite{ross2013flight} and teaching robots by demonstration \cite{englert2013model}. 
In our work, we imitate optimal control actions of a model predictive controller with perfect future information. This relates to explicit model predictive control~\cite{Alessio2009,AKESSON2005323}, but automatically selects the relevant states at which to approximate the optimal policy.

%% file: sections/2-problem.tex

\section{Problem Statement}
\label{sec:problem-statement}

We focus on a PV power plant with ramp rate limitations that uses a battery to compensate sudden shortfalls and sudden overproduction due to cloud movement. As mentioned in the introduction, the lifetime of the battery is increased if its energy throughput,
\begin{equation}
\int |u(t) - s(t)|\, dt ,
\label{eq:battery_throughput}
\end{equation}
is minimized. Here $u(t)$ is the present PV plant power output, which is the controller decision variable, and $s(t)$ is the present power harvested from the PV panels. In the following, we use a point irradiance measurement in $W/m^2$ to quantify~$s(t)$, instead of the output of a real PV plant. This quantity is strongly correlated with PV power output, but is more readily available due to reasons of data ownership. Thus the control input~$u(t)$ corresponds to the PV panel area. Our method is applicable to the case of energy measurements without change.

The power output controller has to respect ramp rate limits on~$u(t)$ in both up and down directions.
We assume realistic rate constraints of~\SI{2/3}{\watt\per\square\meter\per\second}. This corresponds, at a maximum solar power of
\SI{1000}{\watt\per\square\meter}, to a ramp rate limit of 4\% per minute. This is a typical value today and slightly stricter than
the first ramp rate limit of 10\% per minute implemented in Puerto Rico~\cite{gevorgian2013review}.
The commonly used state-of-the-art controller for ramp rate limitation is  purely reactive~\cite{de2015control, makibar2017relation}.
It sets~$u(t)=s(t)$ if possible and otherwise tracks~$s(t)$ at the maximum allowed rate. In this paper, this control policy is used as baseline.

To focus on the problem of learning from images, we assume that the battery has infinite charge. Thus, the battery state of charge
is not relevant. This is a minor restriction and our methods can be easily extended to take the state of charge as an additional input to
the control policy.

\subsection{Future Information as Input}
\label{sec:future-input}

\begin{figure}[t]
	\centering
	\includegraphics[scale=1.0]{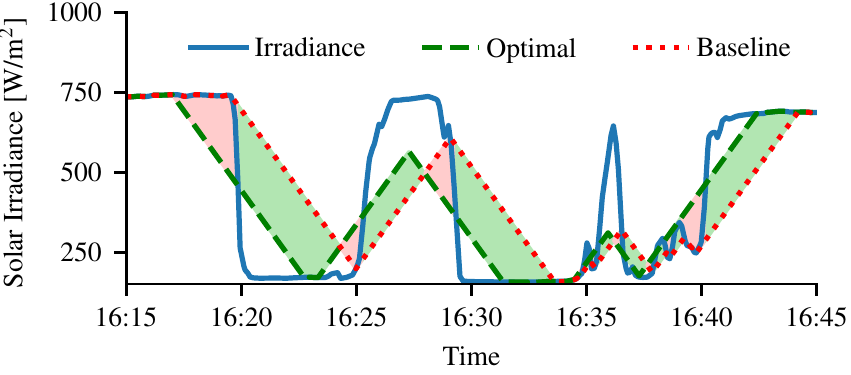}
	\caption{Optimal and baseline policy. The optimal policy minimizes the difference between output and irradiance under the ramp rate constraints. The baseline policy is a reactionary policy that always follows the current irradiance value. Green areas indicate where the optimal policy is better than the baseline.}
	\label{fig:optimal-baseline-example}
\end{figure}

If high-quality predictions of~$s(t)$ were available, we could  minimize~\cref{eq:battery_throughput} directly online using MPC~\cite{camacho2013model}. In this setting, the ramp rate constraint imposes a linear constraint, $|u(t)-u(t + t_s)| \leq t_s \cdot \SI{2/3}{\watt\per\square\meter\per\second}$ for all time steps~$t$, where~$t_s$ is the discretization interval.
Minimizing~\cref{eq:battery_throughput} under this constraint reduces to a linear program, which can be easily solved given perfect future predictions of~$s(t)$. While this approach is not feasible online due to reliance on future data, we can compute the corresponding control actions in hindsight based on our data. We call the resulting policy the \emph{hindsight-optimal controller}. We show an example to compare the behavior of the baseline and hindsight-optimal policy in~\cref{fig:optimal-baseline-example}.
While the baseline policy can only react to changes, the optimal policy acts in a proactive manner and thereby significantly reduces the energy that has to be stored in the battery (green shaded).

Our main controller does not have access to future values of~$s(t)$. Instead, it uses the current irradiance~$s(t)$ together with a sequence of past and current camera images of the sky as input.

%% file: sections/3-background.tex

\section{Background}

In this section, we introduce the imitation learning algorithm and the ResNet neural network architecture that we use as a control policy based on the images of the sky.

\subsection{Imitation Learning}
\label{sec:imitation-learning}

A control policy trained with imitation learning aims to imitate the actions taken by a target policy~$\pi^*$. In our case~$\pi^*$ corresponds to the optimal policy based on the linear program described in~\cref{sec:future-input}. In particular, we are given example trajectories of how the target policy controls the system, ${\mathcal{D} = \{y_t, u_t\}_{t=1}^T}$, which consist of  measurements of the state~$y_t$ and the corresponding control actions taken by~$\pi^*$. The goal is to find a policy~$\pi_\theta$ that performs similarly to the policy~$\pi^*$ when applied on the underlying dynamical system. While the target policy~$\pi^*$ has access to full information, the approximate policy $\pi_\theta$ only observes a partial state or measurements in terms of images.

Written as a supervised learning problem, this corresponds to finding policy parameters that minimize the error on the dataset~$\mathcal{D}$. 
However, even small errors can cause the policy~$\pi_\theta$ to diverge from the optimal trajectory and encounter states for which it has not seen any data in~$\mathcal{D}$. As a consequence, the policy is in general unable to return to the optimal trajectory after an initial error.

One solution to this problem is the dataset aggregation (\daggeralg) algorithm~\cite{ross2011dagger}. It proceeds by repeatedly evaluating the current policy~$\pi_\theta$ on the system and adding the corresponding state measurements together with the corresponding action that the reference policy would have taken to the dataset, $\mathcal{D} \gets \mathcal{D} \cup \{y_t, u_t\}_{t=1}^T$. As we apply this procedure iteratively, the dataset~$\mathcal{D}$ is dominated by state observations that occur under the approximate policy, which ensures that the approximate policy performs well on the real system, rather than only on the trajectories induced by the reference policy. To better leverage the guidance of the optimal policy, \daggeralg{} introduces a modified policy ${\pi_n = \beta_n \pi^* + (1- \beta_n)\pi_\theta}$ that at each iteration~$n$ executes the optimal policy a fraction of the time while collecting the next dataset. This has a positive impact especially at the beginning of training, when the trained policy is still prone to making mistakes and thus may visit states that later become irrelevant as the policy improves. A typical choice for~$\beta_n$ is an exponential decay of the form~$\beta_n=p^{n}$ with~$0\leq p<1$.

\subsection{Neural Networks}
\label{sec:neural_network}

For the problem in~\cref{sec:problem-statement}, we need a policy that can use information in images. The state-of-the-art method for image data are neural networks. These function approximators operate iteratively by repeatedly applying simple linear transformations followed by a nonlinearity to the input~$y$. Starting with the input~$o_1 = y$, at iteration~$l$ the next representation is computed as~$o_{l+1} = f(b_l + W_l o_l)$ based on the result~$o_l$ from the previous iteration. The tuning parameters~$b_l$ and~$W_l$ are learned. Several choices for~$f$ are possible, but a common choice are rectified linear units (ReLU) with~$f(x)=\max(0,x)$. For images, instead of a linear transformation one typically uses convolutional neural networks, which instead convolve~$o_l$ with a learned filter.

A residual network (ResNet~\cite{he2016resnet}) is a particular kind of neural network architecture, which we use in the following. In this architecture, the convolutional layers are grouped into blocks and the input~$o_l$ is considered as an additive feedforward term to the output~$o_{l+1}$. This means that each block learns residuals relative to the identity function. In particular, if a block of layers learns the function $\mathcal F(o_l)$, then the output after the block is $o_{l+1} = o_l + \mathcal F(o_l)$. This particular architecture has achieved state-of-the-art performance in several computer vision competitions.

%% file: sections/4-control.tex

\section{Control from Images}
\label{sec:experimental-setup}

In this section, we show how to use imitation learning together with neural networks in order to solve the control problem defined in~\cref{sec:problem-statement}.

We train a neural network controller with imitation learning based on the optimal MPC control actions in hindsight. We  base our neural network on the ResNet architecture, because it has shown promise in previous work on the same data set~\cite{pothineni2018}. Other previous work has used MPC on irradiance predictions~\cite{habicht2017PVRL} to create control predictions, but did not manage to beat the baseline. In this work, we use MPC on the training data to compute the optimal policy as in~\cref{sec:future-input}, which we use as the target policy for imitation learning.
In particular, we use the \daggeralg{} algorithm from~\cref{sec:imitation-learning} together with a large dataset of past irradiance measurements and sky images.
In the following, we provide the details of the policy architecture, how it is trained, and the dataset below. We evaluate this approach experimentally in~\cref{sec:results}.

\subsection{Dataset and Pre-processing}
\label{sec:dataset}

\begin{figure}[t]
	\centering

	\includegraphics[width=\columnwidth]{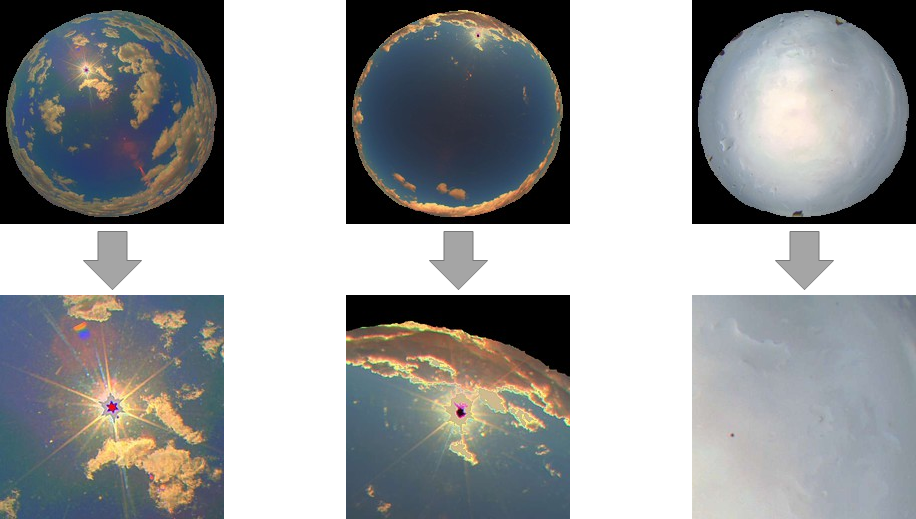}

	\caption{Example images of the sky before and after pre-processing. The images are cropped to an area centered on the sun.}
	\label{fig:dataset-examples}
\end{figure}

We use real-world data collected by ABB Corporate Research from a PV power plant located in Italy. The data consists of whole sky images and various sensor measurements such as irradiance values. The images are post-processed to have a high dynamic range (HDR), which improves the quality as the camera is pointed at the sun. Example images are shown in the top row of \cref{fig:dataset-examples}. Measurements were taken every seven seconds during daytime for 256 days. We assign the days in the dataset at random to the training, validation and test set with a split of 70\%, 15\% and 15\%, respectively.

For neural networks to perform well in practice, preprocessing the images is an important step. In particular, for training it is important to scale all input data to the network to the same value range. Specifically, we scale irradiance values to $[0, 1]$ based on their value range in the dataset and divide each pixel by their maximum possible value, which is 255. This makes it easier to define a robust learning rate for the gradient-descent training algorithm, since all training parameters operate on inputs of similar scale.

To improve performance further, we apply a mask to the images to remove the border of the camera and other background objects. These objects often have coloring issues due to the HDR processing and do not provide relevant information. Next, we center the image on the sun and crop them from $1566\times 1566$ pixels to $448\times 448$. To this end, we compute the sun's position in the sky based on time, camera, and geographic location. This change removes clouds that are far away from the sun and allows us to consider a smaller image at higher resolution than if we included irrelevant parts of the picture. Lastly, we downsample the images by a factor of two to obtain an image of size $224 \times 224$ suitable for training in a neural network. The post-processed images are shown in the bottom row of~\cref{fig:dataset-examples}.

A drawback of the HDR processing is that the colors are not always calibrated consistently. To compensate for this issue, we apply a simple color stabilization method.
We low-pass filter the average intensity of each color channel according to~$\mu_t = 0.9 \mu_{t-1} + 0.1 i_t$, where~$i_t$ denotes the current average intensity of each color channel, ignoring any black pixels of the mask. We then update the color channels in each image by scaling each pixel by~$\mu_t / i_t$. This procedure ensures that color channels are consistent between consecutive images, rather than being corrupted by the HDR processing.

\subsection{Trained Policy}
\label{sec:policy-model}

\begin{figure}[t]
	\centering

	\includegraphics[width=\columnwidth]{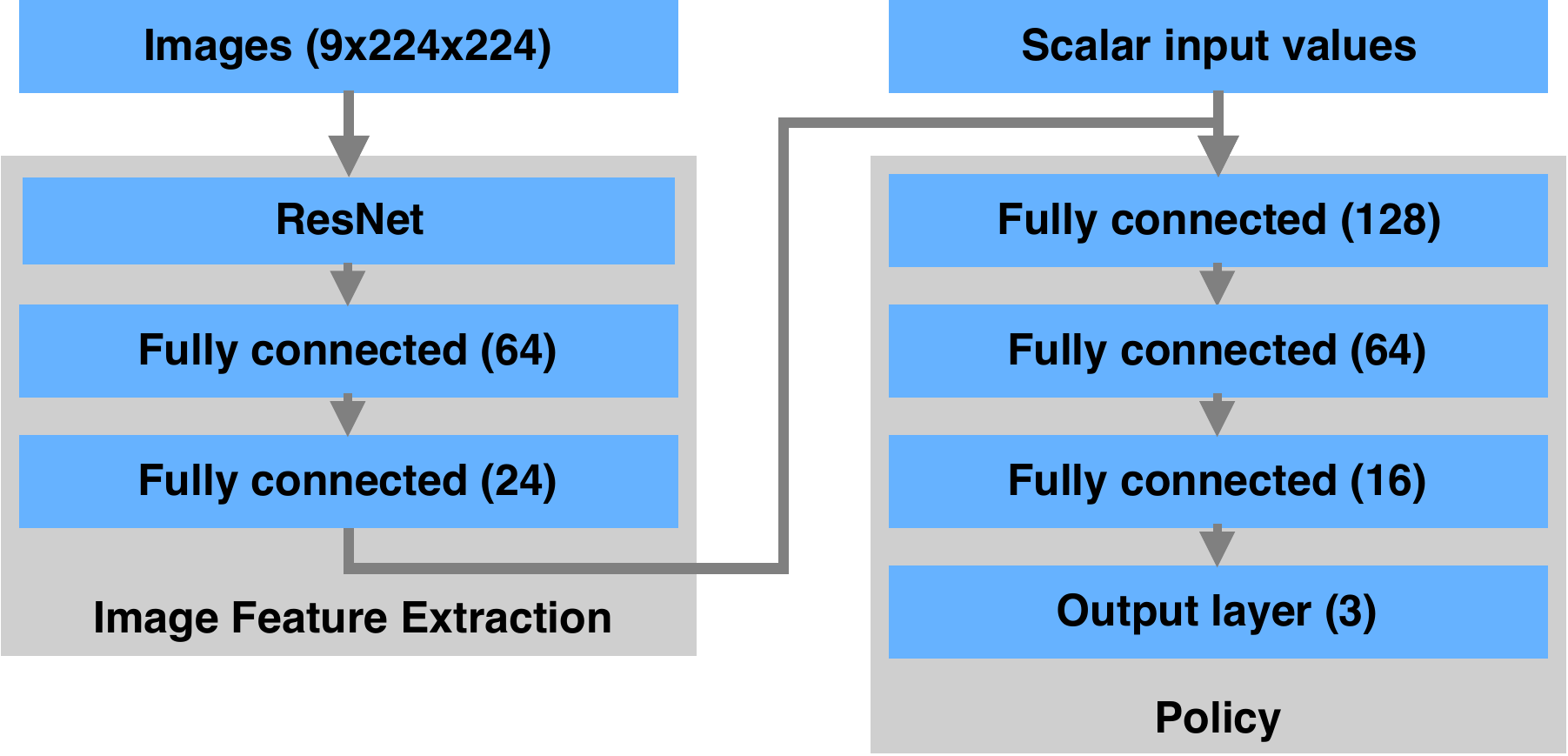}

	\caption{Neural network architecture. The left part of the network uses a ResNet to extract features from the images, while the right hand represents the control policy based on the features and scalar sensor measurements.}
	\label{fig:network-architecture}
\end{figure}

In order to use \daggeralg, we must define a function approximator for the optimal policy. A key observation is that the optimal controller in~\cref{fig:optimal-baseline-example} defaults to~$u(t) = s(t)$ when no environment change is anticipated, but decreases or increases the output at the maximum rate when necessary. This means that we can parameterize our control policy using three discrete actions. In particular, the first action corresponds to the baseline behavior of~$u(t) = s(t)$ within the rate constraints, while the second and third actions correspond to increasing and decreasing the output at the maximum possible rate, ${u(t) = u(t-1) \pm 2/3}$. As a result of this parameterization, the neural network only needs to decide when to deviate from the baseline controller, rather than learning the full controller including basic tracking. This significantly improves performance. Moreover, providing a discrete action set allows us to evaluate a notion of uncertainty of actions, which we explain later.

We structure our neural network to be appropriate for the task at hand, see~\cref{fig:network-architecture}. As an input to the policy we use a sequence of the current image and the images 30~seconds and 1 minute in the past. We include past images to allow the policy to extract cloud movement, which is not contained in a single static image. For processing, each image sequence is passed through an 18 layer ResNet based on convolutions. This choice is motivated by previous work on images in a forecasting task~\cite{pothineni2018}. We then process the output of the ResNet further using two fully connected layers of sizes 64 and 24 to obtain a feature representation for each image sequence. For the last few layers of the network we concatenate the image sequence representation together with the scalar input values such as the current and five previous irradiance measurements and the previous action,~$u(t-1)$. As such, this last part can be interpreted as the core policy based on the ResNet image features. We use three fully connected layers with ReLU activations of sizes 128, 64 and 16.
The output layer uses a softmax function to obtain a normalized probability distribution over the three possible actions. Since we have three discrete actions, we can train the network by minimizing the cross entropy loss between the prediction and the optimal action determined by the precomputed MPC controller.

We use Keras~\cite{chollet2015keras} with the Tensorflow~\cite{tensorflow2015-short} backend to train our neural network. As training the policy including simulations is time consuming, we pretrain the convolutional layers on a simple, supervised prediction task for 50 epochs. In particular, we train the neural network to predict whether future irradiance values in the next 10, 30, 60, and 120 seconds are contained within the ramp rate, higher, or lower. Since this prediction task is a core requirement for the resulting policy, it allows the convolutional layers to already extract useful feature representations from the images. Moreover, since training the supervised policy does not require simulation and recomputing optimal actions for new states, it is significantly cheaper than imitation learning. This reduces the overall training time. After pretraining, we apply \daggeralg{} with~$\beta_n = 0.9^n$, where~$n$ denotes the training epoch, for 50 epochs. New trajectory data is collected every two training epochs, starting after epoch five. This ensures that the network is reasonably trained before the first sampling. For each state, we compute the optimal action using the MPC controller based on perfect information and aggregate the observation-action pair in the training dataset~$\mathcal{D}$. We then optimize the policy by minimizing the prediction error on~$\mathcal{D}$.

\subsection{Confidence Threshold}
\label{sec:confidence-threshold}

\begin{figure}[t]
	\centering
	\includegraphics[scale=1.0]{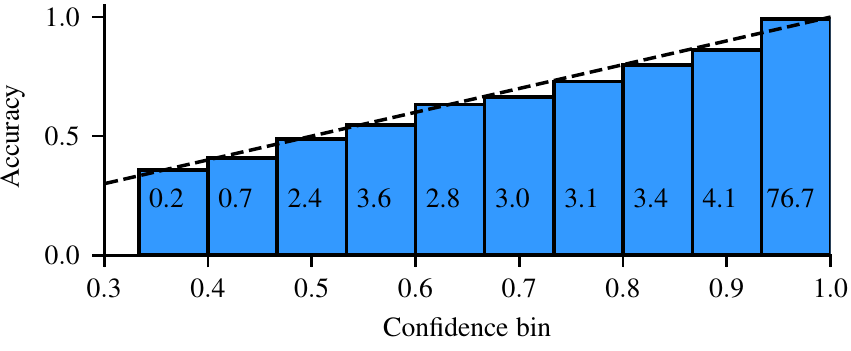}
	\caption{Confidence compared to accuracy. We take the maximum value of the neural network's softmax output as a measure of confidence. The histogram shows the probability of the action being optimal as a function of the neural network confidence prediction, while the number on a bin shows the percentage of samples in that bin. The accuracies are close to the identity function (dashed line), which means that this measure of confidence strongly correlates with the accuracy of the model on the validation set.}
	\label{fig:confidence-calibration}
\end{figure}

The neural network outputs a probability distribution over actions. We use this distribution to estimate the confidence of our model, where a high probability for a single action indicates a high confidence. To evaluate the reliability of these confidence estimates on the real data, we test how strongly the output distribution correlates with the accuracy of our model. To this end, we compute the probability distribution over actions for the states visited during a run on the validation set and bin the actions according to the maximum assigned probability across all three actions. As shown in~\cref{fig:confidence-calibration}, the average accuracy of these bins matches the corresponding confidence of the neural network. While in our case the calibration is already very accurate, in cases when this is not the case this can be corrected using temperature scaling~\cite{guo2017calibration}. Overall the output distribution is meaningful and can be used as a reliable confidence estimate.

In our model, the neural network explicitly decides when to deviation from the baseline controller, see~\cref{sec:policy-model}. The baseline is a very robust policy and diverging from it at the wrong moment can quickly lead to a large loss. Thus, we only want to use the action suggested by the neural network when it has seen a corresponding scenario in the training data and is confident in the predictions. To this end, we only use the neural network action when the confidence estimate of the prediction is above some fixed threshold, and employ the baseline policy otherwise. For real applications, a suitable threshold can be determined, for example, on a validation set that only contains data samples that the model has never seen during training. In~\cref{sec:results}, we evaluate our method on the test data across different confidence thresholds.

%% file: sections/5-results.tex

\section{Experimental Evaluation}
\label{sec:results}

In this section, we evaluate our approach compared to the baseline and optimal policy. Furthermore, to understand the challenges associated
with working with images, we evaluate our model on two intermediate (easier) problems.
The performance of all methods is measured in terms of energy throughput~\cref{eq:battery_throughput} of the battery, see~\cref{sec:problem-statement}.

\textbf{Baseline.}
All methods are evaluated in terms of relative improvement over the baseline reactive controller.

\textbf{Optimal.}
The hindsight-optimal controller is based on perfect future predictions and MPC as described in~\cref{sec:future-input}. Since it is the best possible solution,
we define 100\% improvement over the baseline to be this optimal policy.
Normalizing the relative improvement in this way, makes the following evaluations easier to interpret.

\begin{figure}[t]
	\centering
	\includegraphics{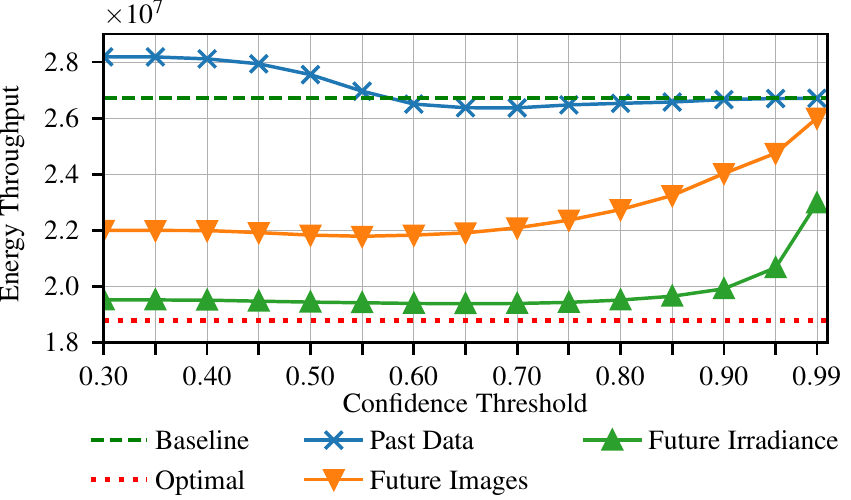}
	\caption{Performance on test set of all models with respect to chosen confidence threshold. The predictions of the models are only executed if the corresponding confidence is above the threshold. For a low threshold, the past data model is more likely to make mistakes on this unseen data. For a higher threshold the performance is very close to the baseline.}
	\label{fig:confidence-threshold-results}
\end{figure}

\textbf{Future Irradiance.}
The first intermediate problem examines whether our neural network is capable of approximating the linear program to minimize~\cref{eq:battery_throughput}.
To this end, we provide irradiance measurements every 15 seconds from the present up to 15 minutes into the future.
In other words, we assume that we are given a perfect forecast of the future irradiance~$s$, which simplifies the problem enormously.
Since the input consists entirely of scalar values, we only use the last three fully connected layers of our architecture.
These are the layers on the right-hand side in~\cref{fig:network-architecture}.

We train these layers with \daggeralg{} for 50 epochs, sampling new data points every second epoch. In~\cref{fig:confidence-threshold-results}
we evaluate the performance on the test set for multiple confidence thresholds, as defined in~\cref{sec:confidence-threshold}.
For a high confidence threshold, only actions with a high confidence estimate are executed and thus the model is closer to the baseline in terms of behavior. For a low confidence threshold the model might make more mistakes.
The best performance is achieved with a threshold of 0.65 resulting in an  improvement of 92\% over the baseline, close to the optimal policy.
This shows that the network is able to approximate the optimal policy when provided with a perfect irradiance forecast. The remaining
approximation errors can be controlled with the confidence threshold.

\textbf{Future Images.}
The second intermediate problem explores if our network is able to extract useful information from images.
We provide a sequence of images of the current and future time steps as input. Specifically, we provide the current image and
images which are 30, 60, 120, 300, and 600 seconds in the future. Since we work with images, we use the full neural network architecture
shown in~\cref{fig:network-architecture}. This experiment evaluates how easy it is to obtain irradiance estimates from images,
and whether the neural network is capable of predicting movement given future images.

We train this model as described in~\cref{sec:policy-model}, pre-training the convolutional layers of the ResNet and afterwards applying
\daggeralg{} for 50 epochs. After training, we evaluate the model on the test set. At a confidence threshold of 0.55, it achieves an improvement
over the baseline of 62\%. This shows that the network is able to extract useful information, such as irradiance estimates, from future images.
However, even with future images as input, the model is no longer able to achieve a near-optimal solution.

\begin{table}[t]
	\caption{Performance overview}
	\label{tab:performance-overview}
	\begin{center}
		\begin{tabular}{lrr}
			Experiment 	& 	Validation $(10^7)$ 	& Test $(10^7)$\\
			\midrule
			Optimal 	& 	100\% (1.448)			& 100\% (1.880)\\
			Future Irradiance 	& 91.62\% (1.497)	& 92.37\% (1.940)  \\
			Future Images 		& 58.62\% (1.690) 	& 62.14\% (2.179)\\
			Past Data 	& 2.62\% (2.018) 		& 4.35\% (2.637) \\
			Baseline  	&  	0\% (2.033)				& 0\% (2.671)\\
		\end{tabular}
	\end{center}
\end{table}

\textbf{Past Data.}
Lastly, we evaluate the performance on the original problem, where the policy only has access to past and present irradiance data and images.
This is a significantly more challenging task compared to the two previous ones, as we have to predict future cloud movement only based on past data.

\begin{figure}[t]
	\centering
	\includegraphics{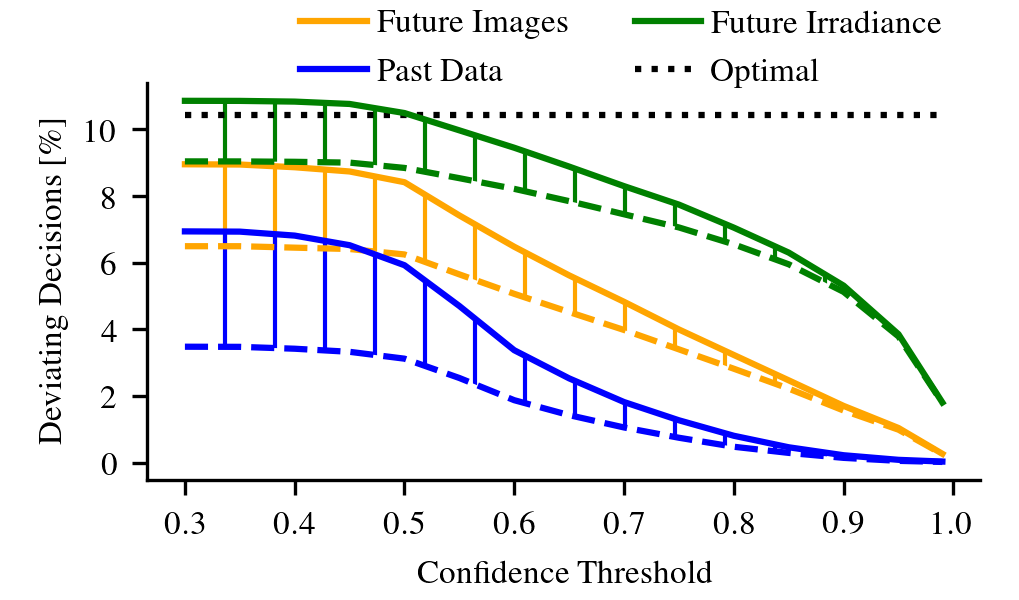}
	\caption{Percentage of actions deviating from baseline with respect confidence threshold. For each model, the upper solid line represents the total percentage of actions deviating from the baseline controller. The lower dashed line shows the percentage of good deviating decisions, which are either equal to the optimal action or lead the policy closer to the optimal trajectory. The hatched area represents unfavorable decisions that increase the loss. The optimal policy itself deviates in about 10\% of the cases from the baseline.}
	\label{fig:deviating-actions}
\end{figure}

We train our model again as described in~\cref{sec:policy-model} but this time the input only consists of past and present data.
On the test set, the best results are achieved with a confidence threshold of 0.7. With only past and present data as input,
the model is able to achieve an improvement of 4.35\% over the baseline.
This shows that while it is possible to improve the control over the baseline,
it is much harder to predict the optimal action without a perfect irradiance forecast or any other information about the future.
This is why choosing the correct confidence threshold is crucial to achieve any improvement. In~\cref{fig:deviating-actions} we show the effects of different confidence thresholds in detail. Specifically, we show how frequently each model deviates from the baseline controller and how often it is a good decision. For a lower confidence threshold, all models take more deviating actions. However, for the model based on past data, the actions are proportionally more often bad than when a higher confidence threshold is chosen. This explains, why the past data model is worse than the baseline for confidence thresholds below~0.6.

\begin{figure}[t]
	\centering
	\includegraphics{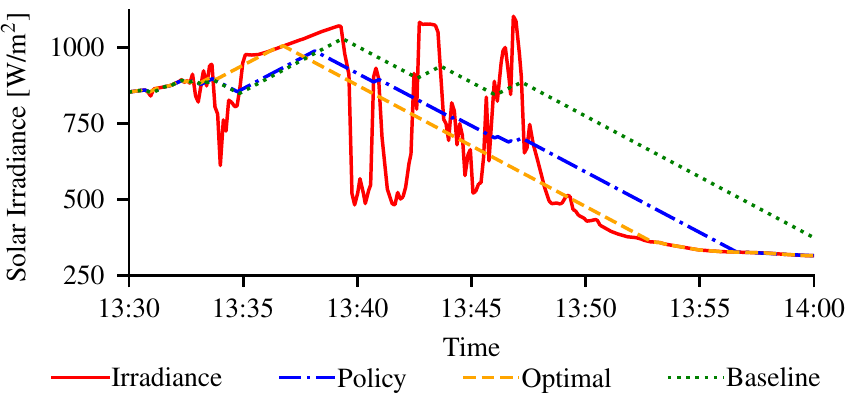}
	\caption{Example episode of the past data model compared to the baseline and optimal policy. Our trained policy is able to anticipate the drop in irradiance around 13:40. Furthermore, it continues to decrease the output at 13:43 when a small gap in the clouds creates a small spike in irradiance. In this way, our policy is able to stay close to the optimal policy.}
	\label{fig:policy-example}
\end{figure}

An example episode of our trained policy compared to the baseline and the optimal policy is shown in \cref{fig:policy-example}. This episode exemplifies how our policy is able to improve over the baseline, by first correctly predicting when a drop in irradiance will occur due to clouds and then continue to decrease the output even when short gaps in the clouds create a short spike in irradiance.

An overview of the performance of all methods is shown in~\cref{tab:performance-overview}, where we also show the performance on the validation set. The absolute value of the energy throughput is displayed in parenthesis. For the setup of ramp rate limitation, this is the first time that the short-term-future information contained in sky images is extracted in a way to demonstrate a practical value.
This method can also be of value for other applications. In microgrids, diesel generators are often used as backup power source. If the microgrid also contains PV systems, then the method presented in this paper could improve the decision making process on whether or not the generators should be turned on.
Moreover, when considering the problem of managing the state of charge of batteries at PV power plants, this method especially together with the confidence threshold could improve the management of the state of charge by preventing unnecessary actions.

%% file: sections/6-conclusions.tex

\section{Conclusions}

In this work, we presented a deep imitation learning approach to optimize the ramp rate control of PV power plants
equipped with a battery.
We showed that a neural network policy, with sky images and irradiance sensor measurements as input, is able to
reduce the energy throughput of the battery compared to a baseline reactive controller.
The following three key factors enable this achievement: (1)~imitation learning towards a hindsight-optimal policy,
(2)~deep neural networks (ResNet) for the image processing, and (3)~uncertainty quantification in order to avoid low-confidence, high-risk actions.